\documentclass[journal]{IEEEtran}
\ifCLASSINFOpdf
  \usepackage[pdftex]{graphicx}
  \graphicspath{{../pdf/}{../jpeg/}}
  \DeclareGraphicsExtensions{.pdf,.jpeg,.png}
\else
  \usepackage[dvips]{graphicx}
  \graphicspath{{../eps/}}
  \DeclareGraphicsExtensions{.eps}
\fi
\usepackage{multirow}
\usepackage{lineno,hyperref}
\modulolinenumbers[5]

\usepackage{booktabs}

\usepackage{amsmath}
\usepackage{algorithmic}
\ifCLASSOPTIONcompsoc
  \usepackage[caption=false,font=normalsize,labelfont=sf,textfont=sf]{subfig}
\else
  \usepackage[caption=false,font=footnotesize]{subfig}
\fi

\usepackage{stfloats}

\usepackage{bm}
\usepackage{algorithm}

\begin{document}
%
\title{MGML: Multi-Granularity Multi-Level Feature Ensemble Network for Remote Sensing Scene Classification}
%
%
%

\author{Qi Zhao ~\IEEEmembership{Member,~IEEE}, Shuchang~Lyu, Yuewen Li, Yujing Ma, Lijiang Chen
\thanks{Qi Zhao is with the Department of Electronics and Information Engineering, Beihang University, Beijing, 100191, China, e-mail: zhaoqi@buaa.edu.cn.}
\thanks{Shuchang Lyu is with the Department of Electronics and Information Engineering, Beihang University, Beijing, 100191, China, e-mail: lyushuchang@buaa.edu.cn.}
\thanks{Yuewen Li is with the Department of Electronics and Information Engineering, Beihang University, Beijing, 100191, China, e-mail: syliyuewen@buaa.edu.cn.}
\thanks{Yujing Ma is with the Department of Electronics and Information Engineering, Beihang University, Beijing, 100191, China, e-mail: zy1902407@buaa.edu.cn.}
\thanks{Lijiang Chen is with the Department of Electronics and Information Engineering, Beihang University, Beijing, 100191, China, e-mail: chenlijiang@buaa.edu.cn.}
}

%
%

\markboth{Journal of \LaTeX\ Class Files,~Vol.~14, No.~8, August~2015}%
{Shell \MakeLowercase{\textit{et al.}}: Bare Demo of IEEEtran.cls for IEEE Journals}
%



\maketitle

\begin{abstract}
Remote sensing (RS) scene classification is a challenging task to predict scene categories of RS images. RS images have two main characters: large intra-class variance caused by large resolution variance and confusing information from large geographic covering area. To ease the negative influence from the above two characters. We propose a Multi-granularity Multi-Level Feature Ensemble Network (MGML-FENet) to efficiently tackle RS scene classification task in this paper. Specifically, we propose Multi-granularity Multi-Level Feature Fusion Branch (MGML-FFB) to extract multi-granularity features in different levels of network by channel-separate feature generator (CS-FG). To avoid the interference from confusing information, we propose Multi-granularity Multi-Level Feature Ensemble Module (MGML-FEM) which can provide diverse predictions by full-channel feature generator (FC-FG). Compared to previous methods, our proposed networks have ability to use structure information and abundant fine-grained features. Furthermore, through ensemble learning method, our proposed MGML-FENets can obtain more convincing final predictions. Extensive classification experiments on multiple RS datasets (AID, NWPU-RESISC45, UC-Merced and VGoogle) demonstrate that our proposed networks achieve better performance than previous state-of-the-art (SOTA) networks. The visualization analysis also shows the good interpretability of MGML-FENet.
\end{abstract}


%
\IEEEpeerreviewmaketitle

\section{Introduction}
%
%
%
%
\IEEEPARstart{R}{emote} Sensing (RS) technology has been widely used in many practical applications, such as RS scene classification~\cite{AID, UC-Merced, NWPU, RSI-CB},  RS object detection~\cite{DOTA, HRSC}, RS semantic segmentation~\cite{Inria, UAVID} and RS change detection\cite{Airchange}. Among the above applications, RS scene classification is a hot topic, which aims to classify RS scene images into different categories.
\begin{figure}[!t]
\centering
\includegraphics[width=1.0\linewidth]{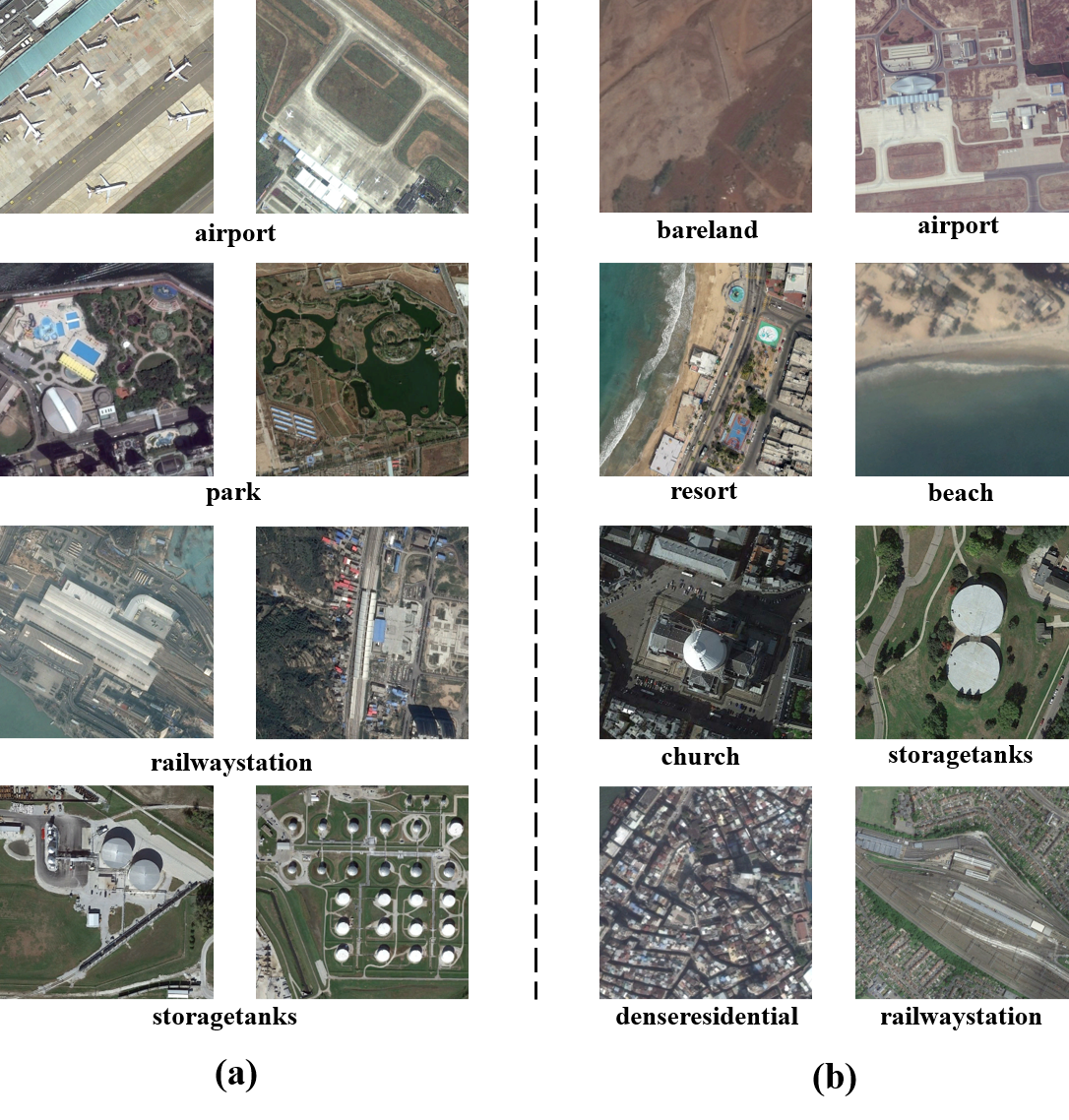}
\caption{Intuitive cases to explain the two main problems in RS scene classification: Figure (a) shows the intra-class variance in RS images. Figure (b) shows the redundant and confusing information in RS images.}
\label{Fig1}
\end{figure}
\par Recent years have witnessed significant progress in various computer vision tasks using deep convolutional neural networks (DCNNs)~\cite{AlexNet, ResNet, RCNN, SPPnet, Faster_RCNN, FCN, XiaoGe}. In some image classification tasks such as scene classification~\cite{SUN, place}, object classification~\cite{ImageNet, CIFAR} and medical image classification~\cite{medicalimg}, DCNNs have shown strong performance by extracting multi-level features with hierarchical architecture~\cite{AlexNet, VGG, ResNet, Identity, SENet, SKNet, Effi, MV3}. Basically, DCNNs efficiently encode each image into a classification probability vector which contains global feature. However, directly using DCNNs to tackle RS scene classification task has two main problems. The first problem is the large intra-class variance caused by resolution variance of RS images (e.g. The image resolution of AID dataset ranges from 0.5$\sim$8 meters~\cite{AID}.), which is intuitively illustrated by some cases in Fig.\ref{Fig1}(a). The second problem is that RS images always contain confusing information because of covering large geographic area. As shown in Fig.\ref{Fig1}(b), confusing information will reduce the inter-class distance. E.g. The inshore ``resort'' is similar to ``beach'' and the ``railwaystation'' built around residential has close character to ``denseresidential''.
\par To address the above two problems, we propose two intuitive assumptions as theory instruction of our method. First, besides global features, fine-grained features are also helpful to RS scene classification. E.g. We can easily recognize ``airport'' if we see planes in RS images. Second, RS images contain latent semantic structural information which can be explored without using detailed annotations like bounding boxes or pixel-level annotations. As shown in third row of Fig.\ref{Fig1}(b), if we want to distinguish ``church'' from ``storagetanks'' , we can't only focus on the center white tower. We need more structural information like ``tower + surroundings'' to make judgement.
\par
\begin{figure}[!t]
\centering
\includegraphics[width=0.9\linewidth]{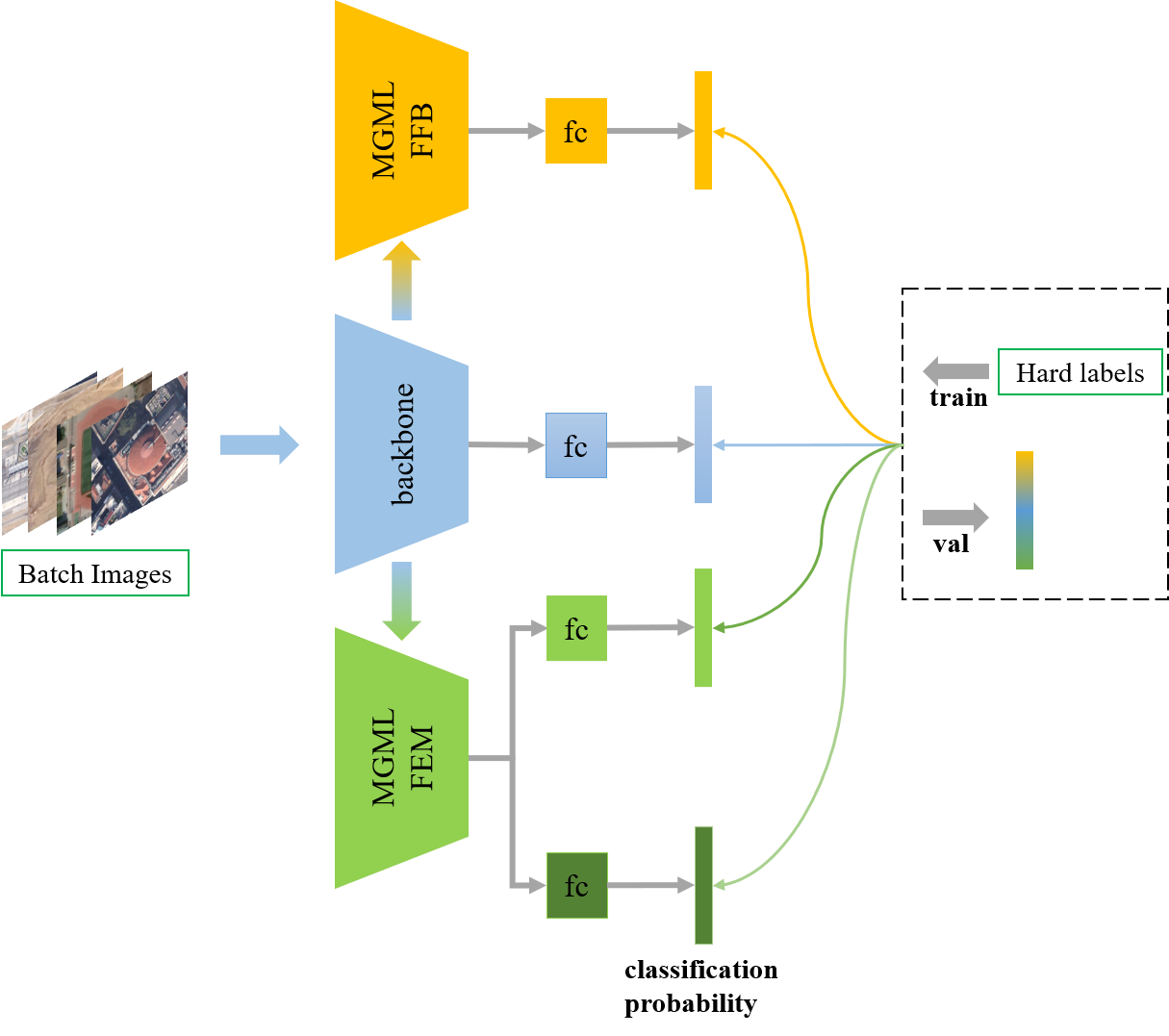}
\caption{The Overview of MGML-FENet architecture. ``MGML-FFB'' denotes multi-granularity multi-level feature fusion branch. ``MGML-FEM'' denotes multi-granularity multi-level feature ensemble module. And ``fc'' denotes fully-connected layers.}
\label{Fig2}
\end{figure}

 Based on the above assumptions, we propose a novel Multi-Granularity Multi-Level Feature Ensemble Network (MGML-FENet) to tackle the RS scene classification task. Specifically, we design multi-granularity multi-level feature fusion branch (MGML-FFB) to explore fine-grained features by forcing the network to focus on a cluster of local feature patches at each level of network. In this branch, we mainly extract aggregated features containing structural information. Furthermore, we propose multi-granularity multi-level feature ensemble module (MGML-FEM) to fuse different high-level multi-granularity features which share similar receptive fields but different resolution. The overview of MGML-FENet is shown in Fig.\ref{Fig2}.
\par In MGML-FENet, MGML-FFB explores multi-granularity multi-level features and utilizes fine-grained features to reduce adverse effects from large intra-class variance. Specifically, we use channel-separate feature generator (CS-FG) to reconstruct feature maps. The original feature map is first cropped into several patches. Each feature patch contains a small group of channels which are split from original feature map. Then, all feature patches are concatenated together to form a new feature map. MGML-FEM utilizes high-level features with structural information to avoid the confusing information interference. In this module, we propose full-channel feature generator (FC-FG) to generate predictions. The first cropping operation on original feature map is the same as CS-FG. Then through global average pooling and concatenation, the new feature vector is created and fed into the classifier at the end of network.
\par To verify the effectiveness of proposed network, we conduct extensive experiments using VGG16~\cite{VGG}, ResNet34~\cite{ResNet} and DenseNet121~\cite{densenet} as baseline models on multiple benchmark datasets (AID~\cite{AID}, NWPU-RESISC45~\cite{NWPU},  UC-Merced~\cite{UC-Merced}) and VGoogle~\cite{VGoogle}. Compared to previous methods, MGML-FENets performs better and achieve new SOTA results.
\par Our main contributions are listed as follows:
\begin{itemize}
\item We propose an efficient multi-granularity multi-level feature ensemble network in RS scene classification to solve the large intra-class variance problem.
\item We derive channel-separate feature generator and full-channel feature generator to extract structural information of RS images which can help solve the confusing information problem.
\item We integrate all features together and construct an end-to-end ensemble networks which achieve better classification results than previous SOTA networks on different benchmark datasets.
\end{itemize}

\section{Related Works}

\subsection{Remote Sensing Scene Classification}
\par In recent years, researchers have introduced many notable methods for RS scene classification. These methods can generally be divided into two types: traditional handcrafted feature-based method and DCNNs based method.
\par Handcrafted feature-based methods always use some notable feature descriptors. \cite{UC-Merced, 2-D_BOVW, BoVW_RS, BoVW_study} investigate bag-of-visual-words (BoVW) approaches for high resolution land-use image classification task. Scale-invariant feature transform (SIFT)~\cite{SIFT} and Histogram of gradient (HoG), two classical feature descriptors, are widely applied in RS scene classification field~\cite{SS_SIFT, SIFT_2, HoG_RS}.
\par Compared to traditional handcrafted feature-based method, Deep convolutional neural networks have better feature representation ability. Recently, DCNNs have achieved great success in RS scene classification task. \cite{DL_RS1, DL_RS2} apply DCNNs to extract features of RS images and further explore its generalization potential to obtain better performance. In addition, some methods integrate attention mechanism into DCNNs to gain more subordinate-level feature only with the guidance of global-level annotations~\cite{att_RS1, att_RS2}. To tackle the inter-class similarity issue and large intra-class variance issue, second order information are efficiently applied in RS scene classification task ~\cite{SCCov, MGCAP}, which receive excellent performance. More recently, Li et al. propose a notable architecture KFBNet to extract more compact global features with the guidance of key local regions~\cite{KFB} which is now the SOTA method. In this paper, we will mainly compare our results with ~\cite{SCCov, MGCAP, KFB}.
\subsection{Multi-Granularity Features Extraction Methods}
\par In some classification tasks like~\cite{CUB200, fg_plane}, the large inter-class similarity will result in rapid performance decline of DCNNs. To better solve this problem, many fine-grained feature extraction methods are proposed~\cite{fg_ex1, fg_ex2}. However, in most cases, only global annotations are provided, which means finding fine-grained features become difficult because of lacking semantic-level annotations. Therefore, multi-granularity feature extraction methods are applied to enhance the region-based feature representation ability of DCNNs~\cite{MGCAP, RACNN, MACNN, 2level, MG_FG}. Inspired by the above methods, we adopt multi-granularity feature extraction in our method to tackle RS images.
\par Ensemble learning based methods offer another perspective to extract multi-granularity features by designing multi-subnets with different structures. \cite{ensem_1} directly uses several CNNs to create different classification results which are then fused via occupation probability. \cite{ensem_2} introduces a learning system which learns specific features from specific deep sub-CNNs. \cite{ensem_3} adopts an ensemble extreme learning machine (EELM) classifier for RS scene classification with generalization superiority and low computational cost. Learning from above ensemble learning based methods, we adopt ensemble learning method in our architecture to integrate multi-granularity and multi-level features.
\begin{figure*}
  \centering
  \includegraphics[width=1.0\linewidth]{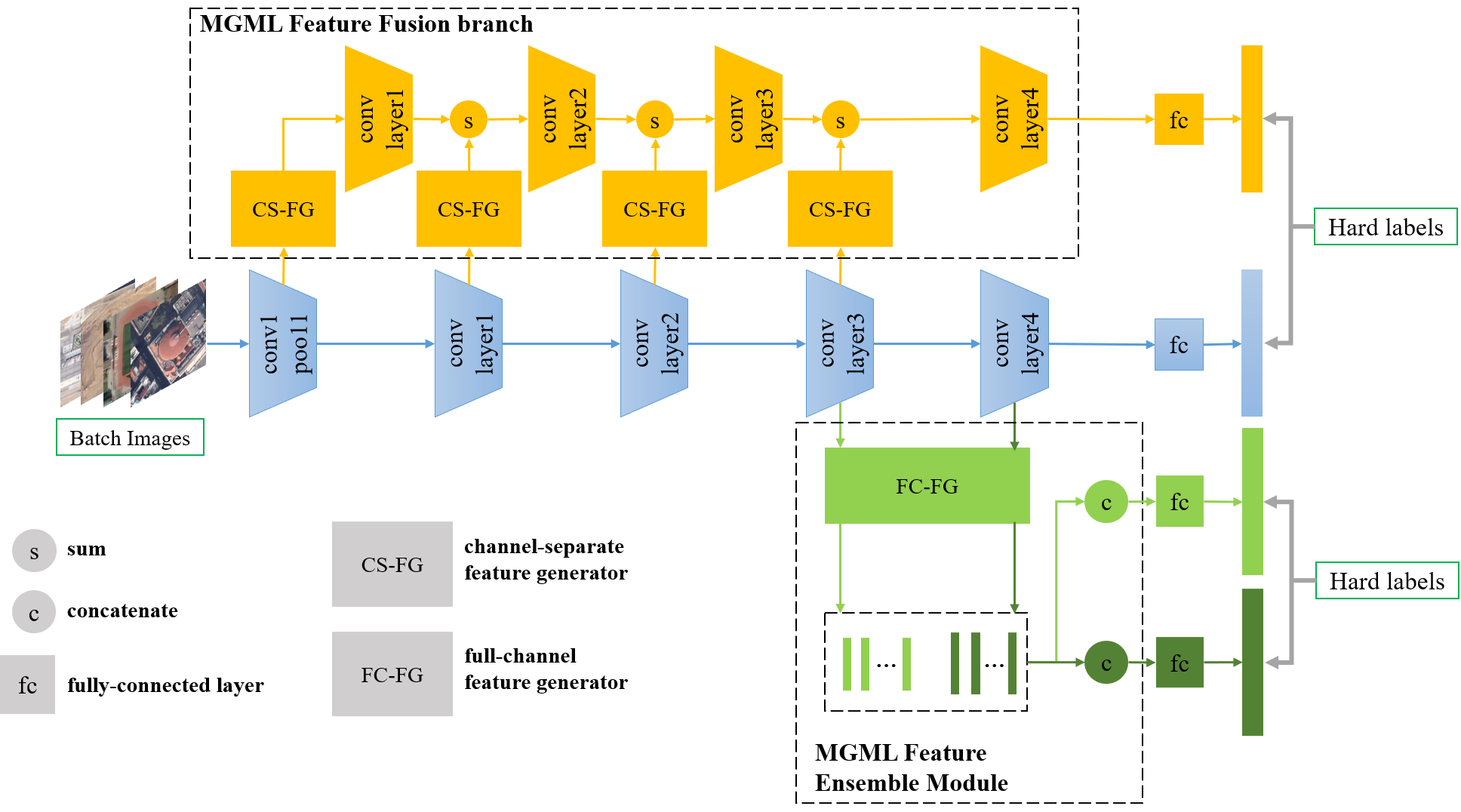}
  \caption{The structure of MGML-FENet. MGML-FENet consists of three parts: main branch (blue), MGML Feature Fusion branch (orange) and MGML Feature Ensemble Module (green). Main branch is ImageNet-pretrained baseline structure (ResNet34/VGG16/DenseNet121). MGML-FFB consists of CS-FG and ``conv layers''. MGML-FEM consists of FC-FG. The ``conv layer'' is basic convolutional blocks. Detailed structure is shown in Tab.\ref{tab1}.}
\label{Fig3}
\end{figure*}
\subsection{Feature Fusion Methods in RS scene classification}
\par To reduce the harm from resolution variance of images, many researchers employ feature fusion method and obtain better performance. Liu et al.~\cite{msCNN} propose a multi-scale CNN (MCNN) framework containing fixed-scale net and a varied-scale net to solve the scale variation of the objects in remote sensing imagery. Zeng et al.~\cite{glbandloc} design a two-branch architecture to integrate global-context and local-object features. ~\cite{ff_1} presents a fusion method to fuse multi-layer features from pretrained CNN models for RS scene classification. In this paper, we also focus on feature fusion method to tackle features which have different granularity, localization and region scales.

\section{Proposed Method}
\par In RS scene classification task, only extracting global feature of RS images can work well in most cases. For the purpose to further improve the performance, we integrate global feature and multi-granularity multi-level features together. Therefore, we propose MGML-FENet to tackle RS scene image classification task. As shown in Fig.\ref{Fig3}, The batch images are first fed into ``conv pool'' (``Conv1''and``Pool1'' in Tab.\ref{tab1}). Then, the output feature map then passes through the four ``conv layers''(``Layer1$\sim$4'' in Tab.\ref{tab1}) and finally generate the final classification probability vector in main branch. At each level of main branch, the feature map are reconstructed by CS-FG and fused with former feature map in MGML-FFB. MGML-FFB offers another classification probability vector. Specifically, the ``conv layers'' in MGML-FFB and main branch use the same structure but do not share parameters, which means more parameters and computation costs are introduced. CS-FG extracts local feature patches and construct new feature map. Compared to original feature map, the new feature map has same channel but smaller scale which eases the computation increase. Output feature maps of the last two main branch layers are served as input to MGML-FEM and generate two classification probability vectors from different levels of networks. Different from MGML-FFB, MGML-FEM brings in few extra parameters and computation.
\par During training, each branch is trained using cross-entropy loss with different weights. During validation, the final classification probability vector of each branch are fused together to vote for the final classification result.

\subsection{Main Branch}
\par In RS images, global feature contains important high-level feature. To extract global feature, we employ main branch in MGML-FENet. As shown in Fig.\ref{Fig3} and Tab.\ref{tab1}, main branch has the same structure as baseline models (VGG16, ResNet34, DenseNet121). In main branch, we denote ``conv1 pool1'' as $f_{0}(\cdot)$ and ``conv layer1$\sim$4''as $f_{1}(\cdot) \sim f_{4}(\cdot)$. The feature map at each level of main branch can be calculated as Eq.\ref{eq1}.
\begin{equation}
    \bm{F_{i}} = f_{i}(\bm{F_{i-1}}), \quad \quad F_{-1} = \textbf{X}
    \label{eq1}
\end{equation}
Where $\bm{F_{i}}$ is the output feature map of $f_{i}$, $\bm{F_{i-1}}$ is feature map from former layer. When $i=0$, the feature map $\bm{F_{-1}}$ is the input image $\textbf{X}$.
\par In addition, we demote the fully-connected layer as $f_{mb}(\cdot)$. The final class-probability prediction ($\bm{P_{mb}}$) is calculated as Eq.\ref{eq2}.
\begin{equation}
    \bm{P_{mb}} = f_{mb}(\bm{F_{4}})
    \label{eq2}
\end{equation}

\begin{figure*}[!t]
  \centering
  \includegraphics[width=1.0\linewidth]{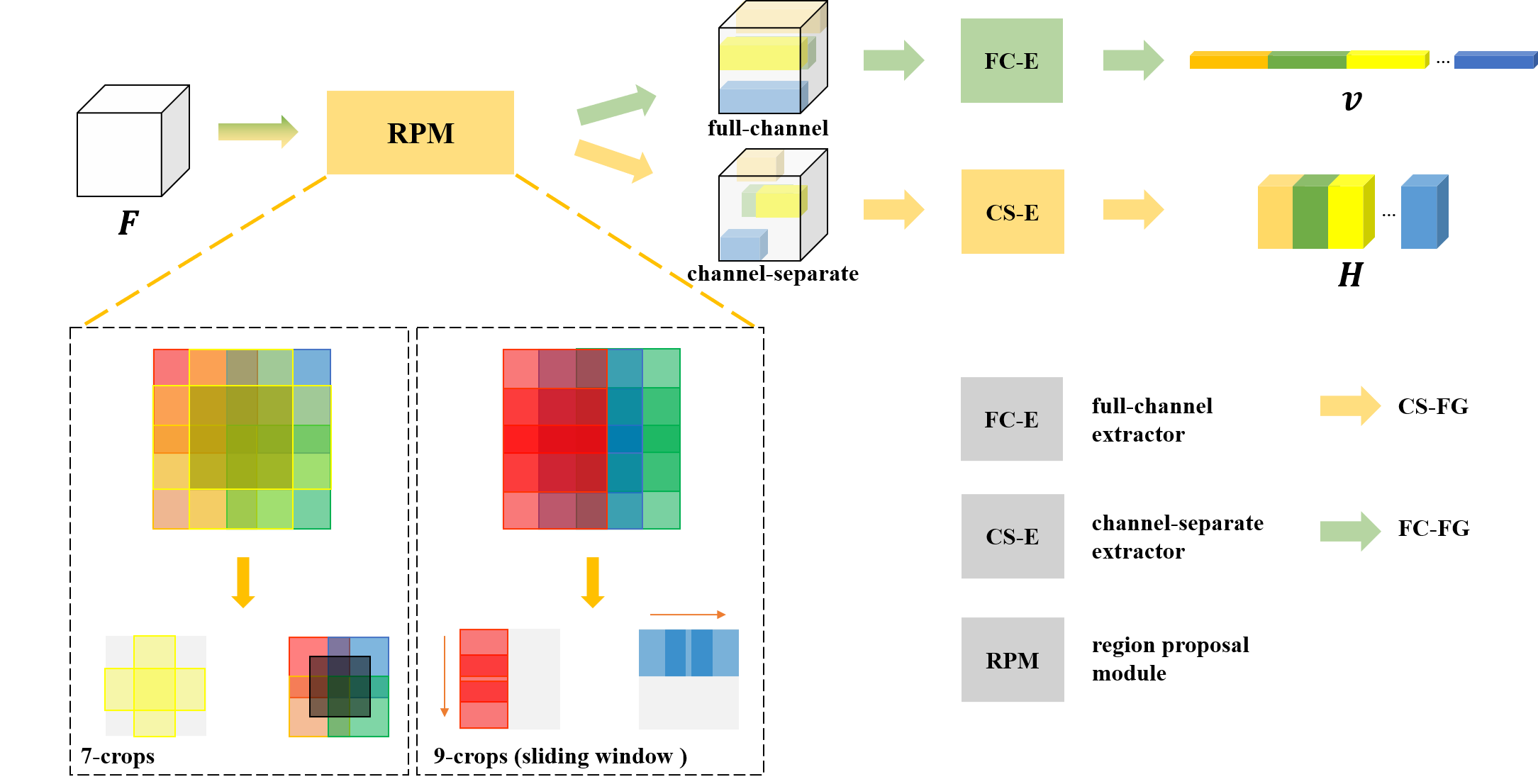}
  \caption{The structure of CS-FG and FC-FG modules. These two modules use the same region proposal module (RPM) to generate several feature patches. After RPM, CS-FG and FC-FG respectively uses channel-separate extractor (CS-E) and full-channel extractor (FC-E) to reconstruct feature patches. Both the two modules takes output feature maps of main branch ($\bm{F}$:$\bm{F_{0}} \sim \bm{F_{3}}$) as input. Output feature maps of CS-FG and FC-FG are $\bm{H}$:$\bm{H_{0}} \sim \bm{H_{3}}$ and $\bm{v}$:$\bm{v_{3}} \sim \bm{v_{4}}$ respectively.}
\label{Fig4}
\end{figure*}

\subsection{MGML Feature Fusion Branch}
\subsubsection{overview of MGML-FFB}
\par To solve large intra-class variance problem, we design multi-granularity multi-level feature fusion branch to utilize fine-grained features in different levels of networks. The structure of MGML-FFB is shown in Fig.\ref{Fig3}. One feature map output from a specific ``conv layer'' of main branch is first fed into CS-FG to generate channel-separate feature map. Next, a ``conv layer'' of MGML-FFB is followed to represent channel-separate feature map and the output feature map is used to fused with the next stage channel-separate feature map.
\par If we respectively denote ``CS-FG'' and ``conv layer'' in MGML-FFB at each level as $h_{i}(\cdot)$ and $g_{i}(\cdot)$, the output feature map ($\bm{G_{i}}$) at each level of MGML-FFB can be calculated as Eq.\ref{eq3} and Eq.\ref{eq4}.
\begin{equation}
    \bm{G_{i+1}} = h_{i+1}(\bm{F_{i+1}}) + g_{i}(\bm{G_{i}}), \quad \quad i=0, 1, 2
    \label{eq3}
\end{equation}
\begin{equation}
    \bm{G_{0}} = h_{0}(\bm{F_{0}}) = h_{0}(f_{0}(\bm{X})), \quad \quad \bm{G_{4}} = g_{3}(\bm{G_{3}})
    \label{eq4}
\end{equation}
\par The final prediction in MGML-FFB can be calculated through another fully-connected layer. The formulation is shown in Eq.\ref{eq5} where ``fc'' layer and the prediction are respectively denoted as $f_{ffb}(\cdot)$ and $\bm{P_{ffb}}$.

\begin{equation}
    \bm{P_{ffb}} = f_{ffb}(\bm{G_{4}})
    \label{eq5}
\end{equation}
\begin{algorithm}[H]
\normalsize
		\caption{7-crop and 9-crop region proposal algorithm}
		\label{Alg1}
		\begin{algorithmic}
			\REQUIRE A feature map \bm{$F_{i}^{C*H*W}$} from main branch, crop scale $\sigma \epsilon (0, 1)$ and stride $s_{H}, s_{W}$ (for 9-crop only).
			\ENSURE {An anchor list: $A_{i} = \{a_{j}\}$. The format of $a_{j}$ is like ``(x1, y1, x2, y2)'' }
			\IF {RPM type is 7-crop}
			\STATE {$a_{0} = (0, 0, W*\sigma, H*\sigma)$}
			\STATE{$a_{1} = (0, H*(1-\sigma), W*\sigma, H)$}
			\STATE{$a_{2} = (W*(1-\sigma), 0, W, H*\sigma)$}
			\STATE{$a_{3} = (W*(1-\sigma), H*(1-\sigma), W, H)$}
			\STATE{$a_{4} = (W*(1-\sigma)/2, H*(1-\sigma)/2, W*(1+\sigma)/2, H*(1+\sigma)/2)$}
			\STATE{$a_{5} = (0, H*(1-\sigma)/2, W, H*(1+\sigma)/2)$}
			\STATE{$a_{6} = (W*(1-\sigma)/2, 0, W*(1+\sigma)/2, H)$}
			\STATE{$A_{i} = \{a_{j}\}, \quad j\epsilon [0, 6]$}
			\ENDIF
			\IF {RPM type is 9-crop}
			\STATE{$(k+1)^2 = 9, \quad count = 0$}
			\STATE{$s_{H} = H * (1-\sigma) / k, \quad s_{W} = W * (1-\sigma) / k$}
			\FOR{$m = 0$, \ldots $k$}
			    \FOR{$n = 0$, \ldots $k$}
			        \STATE{$a_{count} = (m*s_{W}, n*s_{H}, m*s_{W}+W*\sigma, n*s_{H}+H*\sigma)$}
			        \STATE{$count = count + 1$}
			    \ENDFOR
			\ENDFOR
			\STATE{$A_{i} = \{a_{j}\}, \quad j\epsilon [0, (k+1)^2-1]$}
			\ENDIF
		\end{algorithmic}
		
	\end{algorithm}

\subsubsection{CS-FG: Channel-separate feature generator}
\par To utilize fine-grained features and explore the structural information of multi-granularity features, We design CS-FG in MGML-FFB. In each level of MGML-FFB, CS-FG reconstructs original feature by extracting several local feature patches and combining them together. Compared to feature maps in main branch, feature maps in MGML-FFB focus more on local feature rather global feature. Moreover, CS-FG increases the diversity of feature representation which helps a lot on representation RS images. CS-FG is the core module of MGML-FFB. The structure is shown in Fig.\ref{Fig4}. CS-FG consists of region proposal module (RPM) and channel-separate extractor (CS-E).
\par RPM is used to crop original feature maps and generate feature patches. In this paper, we mainly introduce two approaches: 7-crop and 9-crop (sliding windows). In Fig.\ref{Fig4}, it is clear that 7-crop approach extracts seven fix-position patches (left-top, left-bottom, right-top, right-bottom, center, band in middle row, band in middle column) on feature map and 9-crop approach extracts nine fix-position patches using sliding window strategy. In addition, 9-crop approach can be extended to k-crop. In this paper, we set k to 9. The 7-crop and 9-crop region proposal algorithm is shown in Alg.\ref{Alg1}.
\par CS-E is used to extract feature patches on original feature map using anchors $A$, which is generated by RPM (Alg.\ref{Alg1}). And then through recombining feature patches together, the new feature map contains the structural information. As shown in Fig.\ref{Fig4}, feature patches in different locations are concatenated in channel-wise and each feature patch uses separate group of channels. Therefore, when concatenating together, the total channels of new feature map keep unchanged. In CS-E, the input are \bm{$F_{i}^{C*H*W}$} and $A_{i}$, the output are \bm{$H_{i}^{C*\frac{H}{2}*\frac{W}{2}}$}. We introduce the algorithm in Alg.\ref{Alg2}. With channel-separate extractor, the information of different local feature patches are integrated together. Local patches have less spatial information so that only a few group of separate channels are employed. CS-E can maximally utilize the channel-wise information and explore the structural information.
\par In summary, CS-FG consists of RPM and CS-E. In Eq.\ref{eq3}, CS-FG is denoted as $h_{i}(\cdot)$. To express CS-FG in detail, we denote RPM as $h_{i}^{0}(\cdot)$ and CS-E as $h_{i}^{1}(\cdot)$. The detailed expression of CS-FG is in Eq.\ref{eq6} and Eq.\ref{eq7}.

\begin{equation}
    A_{i} = h_{i}^{0}(\bm{F_{i}})
    \label{eq6}
\end{equation}
\begin{equation}
    \bm{H_{i}} = h_{i}(\bm{F_{i}}) = h_{i}^{1}(\bm{F_{i}}; A_{i}) = h_{i}^{1}(\bm{F_{i}};  h_{i}^{0}(\bm{F_{i}}))
    \label{eq7}
\end{equation}

\subsection{MGML Feature Ensemble Module}
\subsubsection{overview of MGML-FEM} To avoid the confusing information interference, we propose MGML feature ensemble module. This module can utilize high-level features with structural information which makes the whole network more robust. Moreover, it provide diverse predictions based on ensemble learning theory to vote for the final classification result. To generate more convincing predictions and make the network train in a reasonable manner, we only apply MGML-FEM in deeper level of network. Because features in shallow layers always contains more low-level basic information. Fig.\ref{Fig3} shows the structure of MGML-FEM.
\begin{algorithm}[H]
\normalsize
		\caption{Channel-separate extractor algorithm}
		\label{Alg2}
		\begin{algorithmic}
			\REQUIRE A feature map \bm{$F_{i}^{C*H*W}$} from main branch, an anchor list: $A_{i} = \{a_{j}\}$. The format of $a_{j}$ is like ``(x1, y1, x2, y2)''. The number of local patches $k$.
			\ENSURE {A feature map \bm{$H_{i}^{C*\frac{H}{2}*\frac{W}{2}}$}.}
			\STATE {Separate channels of input features: $C^{'} = \lfloor C / k \rfloor$}
			\STATE {Extract feature patches:}
			\FOR {$j = 0$, \ldots $k-1$}
			    \IF {$j = k - 1$}
			        \STATE {\bm{$H_{i, j}$} = \bm{$F_{i}$}$[j*C^{'}:C, A_{j}[1]: A_{j}[3], A_{j}[0]: A_{j}[2]]$}
			    \ELSE
			        \STATE{\bm{$H_{i, j}$} = \bm{$F_{i}$}$[j*C^{'}:(j+1)*C^{'}, A_{j}[1]: A_{j}[3], A_{j}[0]: A_{j}[2]]$}
			    \ENDIF
			    \STATE {Downsample feature patches using adaptive pooling (the output size is half of input size): \bm{$H_{i, j}$} = adapool(\bm{$H_{i, j}$})}
			\ENDFOR
			\STATE {Concatenate feature patches: \bm{$H_{i}$} = $[\bm{H_{i, 0}}, \cdots, \bm{H_{i, k-1}]}$}
		\end{algorithmic}
	\end{algorithm}
\par Mathematically, we denote the operation of MGML-FEM as $l(\cdot)$. The output feature vectors ($\bm{v_{i}}$) can be calculated in Eq.\ref{eq8}. In Fig.\ref{Fig3}, it is clear that we only use the feature maps from last two ``conv layers'' of main branch. Towards these two output vectors which have different length, we design two fully-connected layers to generate predictions, which is shown in Eq.\ref{eq9}.
\begin{equation}
    \bm{v_{i}} = l_{i}(\bm{F_{i}}) \quad \quad i = 3, 4
    \label{eq8}
\end{equation}
\begin{equation}
    \bm{P_{fem3}} = f_{fem3}(\bm{v_{3}}), \quad \quad \bm{P_{fem4}} = f_{fem4}(\bm{v_{4}})
    \label{eq9}
\end{equation}
where the fully-connected layers of ``conv layer3'' and ``conv layer4'' are represented as $f_{fem3}$ and $f_{fem4}$ respectively. And the corresponding predictions are represented as \bm{$P_{fem3}$} and \bm{$P_{fem4}$}.
\subsubsection{FC-FG: Full-channel feature generator}
FC-FG is the main part in MGML-FEM. This module mainly extracts high-level features to contribute to the final prediction. As shown in Fig.\ref{Fig4}, FC-FG is formed by RPM and FC-E. RPM in FC-FG is the same as the one in CS-FG. FC-E keeps full-channel information for each feature patches other than uses channel-separate strategy because high-level features need sufficient channel-wise representation. Moreover, FC-E directly uses global average pooling to generate feature vectors because neurons at every pixels of high-level feature have large receptive fields and contain decoupled information. Alg.\ref{Alg3} clearly describes the method of FC-E.
\par To mathematically express FC-FG, we denote FC-E as $l^{'}(\cdot)$. RPM in FC-FG is represented as Eq.\ref{eq6} shows. The detailed expression of FC-FG is listed in Eq.\ref{eq10}.

\begin{equation}
    \bm{v_{i}} = l_{i}(\bm{F_{i}}) = l_{i}^{'}(\bm{F_{i}}; A_{i}) = l_{i}^{'}(\bm{F_{i}};  h_{i}^{0}(\bm{F_{i}}))
    \label{eq10}
\end{equation}

\begin{algorithm}[H]
\normalsize
		\caption{Full-channel extractor algorithm}
		\label{Alg3}
		\begin{algorithmic}
			\REQUIRE A feature map \bm{$F_{i}^{C*H*W}$} from main branch, an anchor list: $A_{i} = \{a_{j}\}$. The format of $a_{j}$ is like ``(x1, y1, x2, y2)''. The number of local patches $k$.
			\ENSURE {A feature vector \bm{$v_{i}^{(Ck)*1}$}.}
			\STATE {Extract feature patches:}
			\FOR {$j = 0$, \ldots $k-1$}
			    \STATE{\bm{$F_{i, j}^{'}$} = \bm{$F_{i}$}$[0: C-1, A_{j}[1]: A_{j}[3], A_{j}[0]: A_{j}[2]]$}
			    \STATE {Downsample feature patches using global average pooling: \bm{$v_{i, j}$} = glbpool(\bm{$F_{i, j}^{'}$})}
			\ENDFOR
			\STATE {Concatenate feature patches: \bm{$v_{i}$} = $[\bm{v_{i, 0}}, \cdots, \bm{v_{i, k-1}]}$}
		\end{algorithmic}
	\end{algorithm}

\subsection{Optimizing MGML-FENet}
\par MGML-FENet models apply conventional cross-entropy loss in every branches during training. To make the network converge well, we allocate each loss a reasonable factor. As shown in Fig.\ref{Fig4}, the whole objective function consists of four cross-entropy losses. We optimize our MGML-FENet by minimize the objective function (Eq.\ref{eq11}).

\begin{equation}
\begin{split}
    L_{obj}(\bm{X}|\bm{Y}) =
    \lambda_{1} * L_{cn}(\bm{P_{mb}}|\bm{Y}) + \lambda_{2} * L_{cn}(\bm{P_{ffb}}|\bm{Y}) + \\
    \lambda_{3} * L_{cn}(\bm{P_{fem3}}|\bm{Y}) +
    \lambda_{4} * L_{cn}(\bm{P_{fem4}}|\bm{Y})
    \label{eq11}
\end{split}
\end{equation}

where $L_{obj}(\bm{X}|\bm{Y})$ and $L_{cn}(\cdot)$ respectively denotes the objective loss and cross entropy loss. $Y$ denotes the hard label. $\lambda_{1} \sim \lambda_{4}$ is four weighted factors to constrain the training intensity of each branch. In this paper, we set ($\lambda_{1}, \lambda_{2}, \lambda_{3}, \lambda_{4}$) as (1, 0.5, 0.2, 0.5) following two main principles. 1. global features can work well in most cases. Therefore, the main branch is supposed to have the highest training intensity. 2. \bm{$P_{fem3}$} outputs from shallower layer so the training intensity should be the lowest.
\par During validation, MGML-FENet employs ensemble learning method, which integrates all predictions to vote for the final result. The final predictions contain diverse information including global information, multi-granularity multi-level information and high-level structural information. Eq.\ref{eq12} calculates the final prediction $\bm{P}$. In addition, MGML-FFB and MGML-FEM in MGML-FENet can easily be dropped from or inserted into main branch as independent parts, which make the whole network flexible.

\begin{equation}
    \bm{P} = \bm{P_{mb}} + \bm{P_{ffb}} + \bm{P_{fem3}} + \bm{P_{fem4}}
    \label{eq12}
\end{equation}

\section{Experiments}
\subsection{Datasets}
\par In this paper, we mainly evaluate our method on four benchmark datasets in RS scene classification task, which include UC Merced~\cite{UC-Merced}, AID~\cite{AID}, NWPU-RESISC45~\cite{NWPU} and VGoogle~\cite{VGoogle}. UC-Merced dataset contains 21 scene categories and total 2100 RGB images with 256 $\times$ 256 pixels. Each category consists of 100 images. All images have same spatial resolution (0.3 meter). AID dataset contains 30 scene categories and total 10000 large scale RGB images with 600 $\times$ 600 pixels. Each category has 220 $\sim$ 420 images. The image spatial resolution varies from 0.5 $\sim$ 8 meter. NWPU-RESISC45 dataset contains 45 scene categories and total 31500 RGB images with 256 $\times$ 256 pixels. Each category consists of 700 images. The image spatial resolution varies from 0.2 $\sim$ 30 meter. VGoogle dataset is constructed by V-RSIR. It's a new large RS scene datasets containing 59404 RGB images and 38 categories. The resolution varies from 0.075 $\sim$ 9.555 meters. There are at least 1500 training samples for each category. Due to the lack of previous results on VGoogle dataset, we compare the classification results between baseline model and MGML-FENet. The classification results on VGoogle are shown in Appendix A.

\subsection{Implementation Details}
\par In this paper, we use ResNet34~\cite{ResNet,Identity}, VGG16~\cite{VGG} and DenseNet121~\cite{densenet} as baseline models to make fair comparison with previous methods. The detailed structure of baseline models are shown in Tab.\ref{tab1}. We select VGG16 as baseline model because many previous methods use VGG16 to extract features. Compared to VGG16, ResNet34 performs better in image classification task using less trainable parameters and FLOPs. Therefore, we also select it as baseline model. As for DenseNet121, \cite{KFB} mainly uses it as baseline model. To make fair comparison, we also choose it as another baseline model.
\par During experiments, we apply fixed training settings for baseline models and our proposed models. First, we use stochastic gradient descent (SGD) with momentum of 0.9 and weight decay of 0.0005. The initial learning rate is set to 0.005 and the mini-batch size is set to 64. The total number of training epochs is 200 and learning rate will be divided by 10 at epoch 90 and 150. For all models, we adopt ImageNet~\cite{ImageNet} pretrained strategy and tune models on RS image datasets. In addition, all models are implemented using Pytorch on NVIDIA GTX 1080ti. Our code will be soon available online.
\begin{table*}[!t]
\begin{center}
\caption{The detailed structure of three baseline models~\cite{ResNet, VGG, densenet}. To clearly show the structure of baseline model, we choose the common-use input image size (224 $\times$ 224) as example. In experiments, the input size maybe different on different datasets. The output vector size is equal to the number of categories. ``Layer1$\sim$4'' denotes four convolutional blocks as shown in Fig.\ref{Fig2}. To lower the number of trainable parameters and computation cost, we modify the fully-connected layer of VGG16 to make it more compact.}
\label{tab1}
\setlength{\tabcolsep}{0.8mm}{
\begin{tabular}{ccccc}
  \cmidrule(r){1-5}
  \multirow{2}{*}{Layers} &
  \multirow{2}{*}{Output size} &
  \multicolumn{3}{c}{baseline models}
  \\ \cmidrule(r){3-5}
  & & VGG16 & ResNet34 & DenseNet121
  \\ \cmidrule(r){1-5}
  Conv1 & 112$\times$112 & conv1-x 2$\times$2 max pool stride=2 & \multicolumn{2}{c}{7$\times$7 conv stride=2}
  \\ \cmidrule(r){1-5}

  Pool1 & 56$\times$56 & - & \multicolumn{2}{c}{3$\times$3 max pool stride=2}
  \\ \cmidrule(r){1-5}

  \multirow{2}{*}{Conv layer1} & \multirow{2}{*}{56$\times$56} & \multirow{2}{*}{conv2-x 2$\times$2 max pool stride=2} & \multirow{2}{*}{$\begin{bmatrix} 3\times3, \quad 64 \\ 3\times3, \quad 64 \end{bmatrix} \times3$}  & \multirow{2}{*}{$\begin{bmatrix} 1\times1 \quad \rm conv \\ 3\times3 \quad \rm conv \end{bmatrix} \times 6$}
  \\ \\ \cmidrule(r){1-5}

  \multirow{3}{*}{Conv layer2} & \multirow{3}{*}{28$\times$28} & \multirow{3}{*}{conv3-x 2$\times$2 max pool stride=2} & \multirow{3}{*}{\shortstack{downsample 2$\times$ \\ $\begin{bmatrix} 3\times3, \quad 128 \\ 3\times3, \quad 128 \end{bmatrix} \times3$}}  & \multirow{3}{*}{\shortstack{transition pool 2$\times$ \\ $\begin{bmatrix} 1\times1 \quad \rm conv \\ 3\times3 \quad \rm conv \end{bmatrix} \times 12$}}
  \\ \\ \\ \cmidrule(r){1-5}

  \multirow{3}{*}{Conv layer3} & \multirow{3}{*}{14$\times$14} & \multirow{3}{*}{conv4-x 2$\times$2 max pool stride=2} & \multirow{3}{*}{\shortstack{downsample 2$\times$ \\ $\begin{bmatrix} 3\times3, \quad 256 \\ 3\times3, \quad 256 \end{bmatrix} \times3$}}  & \multirow{3}{*}{\shortstack{transition pool 2$\times$ \\ $\begin{bmatrix} 1\times1 \quad \rm conv \\ 3\times3 \quad \rm conv \end{bmatrix} \times 24$}}
  \\ \\ \\ \cmidrule(r){1-5}

  \multirow{3}{*}{Conv layer4} & \multirow{3}{*}{7$\times$7} & \multirow{3}{*}{conv5-x 2$\times$2 max pool stride=2} & \multirow{3}{*}{\shortstack{downsample 2$\times$ \\ $\begin{bmatrix} 3\times3, \quad 512 \\ 3\times3, \quad 512 \end{bmatrix} \times3$}}  & \multirow{3}{*}{\shortstack{transition pool 2$\times$ \\ $\begin{bmatrix} 1\times1 \quad \rm conv \\ 3\times3 \quad \rm conv \end{bmatrix} \times 16$}}
  \\ \\ \\ \cmidrule(r){1-5}

  Pool2 & 1$\times$1 & \multicolumn{3}{c}{7$\times$7 global avg pool}
  \\ \cmidrule(r){1-5}

  \multirow{3}{*}{FC} & \multirow{3}{*}{1$\times$1} & \multirow{3}{*}{\shortstack{512$\times$512 \\ 512$\times$512 \\ 512$\times$num\_cls} } & \multirow{3}{*}{512$\times$num\_cls}  & \multirow{3}{*}{1024$\times$num\_cls}
  \\ \\ \\ \cmidrule(r){1-5}
\end{tabular}}
\end{center}
\end{table*}

\subsection{Experimental Results}
\par We conduct extensive experiments to show the performance of MGML-FENet. To evaluate our model, we use overall accuracy  as criterion, which is common-used metric in classification task. Previous methods use different networks as backbone. Therefore, we apply same backbone as previous methods to make fair comparison. To make the results more convincing, we both compare the performance with previous models and baseline models.
\par In RPM of MGML-FENet, we mainly adopt the 7-crop strategy because RS images always contain important information in the middle ``band'' patches according to intuitive observation. We will also compare ``9-crop'' with ``7-crop'' in ablation study.
\subsubsection{Classification on AID dataset}
Following the setting of previous methods on AID dataset, we randomly select 20\% or 50\% data as training data and the rest data are served as testing data. We run every experiments five times to give out the mean and standard deviation of overall accuracy (OA). The comparison results are shown in Tab.\ref{tab2}.
\par If taking VGG16 as backbone, MGML-FENet shows better performance than the SOTA method, KFBNet~\cite{KFB}. Especially when training rate is 50\%, MGML-FENet achieves 97.89\% OA which surpasses KFBNet by 0.7\%. When applying DenseNet121 as backbone, MGML-FENet performs even stronger. It achieves 96.45\% and 98.60\% OA which improves the SOTA accuracy by 0.95\% and 1.2\% when T.R.=20\% and 50\% respectively. In this paper, we introduce ResNet34 as one of the backbone. Because ResNet34 is proven better than VGG16 in image classification field with far less trainable parameters and computation cost. Results in Tab.\ref{tab2} clearly show that MGML-FENet (ResNet34) performs surprisingly better than MGML-FENet (VGG16) and other previous methods.
\subsubsection{Classification on NWPU-RESISC45 dataset}
NWPU-RESISC45 contains more images and categories than AID dataset, so previous methods choose to use 10\% and 20\% images for training. From Tab.\ref{tab2}, MGML-FENet with VGG16 as backbone achieves SOTA results (from 92.95\% to 93.36\%) on 20\% training rate. Also, with backbone DenseNet121, MGML-FENet obtains the best accuracy 95.39\% when T.R.=20\%. Although under the training rate 10\%, MGML-FENet does not obtain SOTA results with VGG16 and DenseNet121, the gap is close (0.14\% and 0.17\%).
\subsubsection{Classification on UC-Merced}
UC-Merced dataset only has 2100 images with 21 categories. The training rate is 80\%, which means only 420 images will be served as val data. Tab.\ref{tab2} shows that KFBNet achieves 99.88\% and 99.76\% classification accuracy respectively using VGG16 and Dense121 as backbone. The results are close to 100\%. Compared to KFBNet, MGML-FENet also achieves high accuracy close to full marks. As mentioned before, we run every experiments five time and calculate the mean and standard deviation of overall accuracy (OA). On VGG16-based model, the five time classification results are 99.76\%, 99.76\%, 99.76\%, 100\% and 99.76\%. On ResNet-based model, we also obtain once 100\% accuracy and four times 99.76\% as the results of VGG16-based model. On DenseNet121-based model, we get one more 100\% compared to above two models. Additionally, 99.76\% accuracy means only one image is recognized as wrong category. From the comparison results, we can observe that our method and KFBNet both reach the ultimate limit on UC-Merced dataset and have obvious advantage against other previous methods.
\begin{table*}
\begin{center}
\caption{Comparison of classification results (\%) on UC-Merced, AID and NWPU-RESISC45 datasets.}
\label{tab2}
\begin{tabular}{cccccc}
  \cmidrule(r){1-6}
  \multirow{2}{*}{Methods(backbone)} &  \multicolumn{2}{c}{AID} & \multicolumn{2}{c}{NWPU-RESISC45} &
  UC-Merced
  \\ \cmidrule(r){2-6}
  {} & T.R.=20\% & T.R.=50\% & T.R.=10\% & T.R.=20\% & T.R.=80\%
  \\ \cmidrule(r){1-6}
  MSCP (VGG16)~\cite{MSCP} & 92.21$\pm$0.17 & 96.56$\pm$0.18 & 88.07$\pm$0.18 & 90.81$\pm$0.13 & 98.40$\pm$0.34
  \\ \cmidrule(r){1-6}
  DCNN (VGG16)~\cite{DCNN} & 90.82$\pm$0.16 & 96.89$\pm$0.10 & 89.22$\pm$0.50 & 91.89$\pm$0.22 & 98.93$\pm$0.10
  \\ \cmidrule(r){1-6}
  RTN (VGG16)~\cite{att_RS1} & 92.44 & - & 89.90 & 92.71 & 98.96
  \\ \cmidrule(r){1-6}
  SCCov (VGG16)~\cite{SCCov} & 93.12$\pm$0.25 & 96.10$\pm$0.16 & 89.30$\pm$0.35 & 92.10$\pm$0.25 & 99.05$\pm$0.25
  \\ \cmidrule(r){1-6}
  MG-CAP (VGG16)~\cite{MGCAP} & 93.34$\pm$0.18 & 96.12$\pm$0.12 &  $ \textbf{90.83}\pm\textbf{0.12}$ & 92.95$\pm$0.13 & 99.0$\pm$0.10
  \\ \cmidrule(r){1-6}
  Hydra (DenseNet121)~\cite{Hydra} & - & - & 92.44$\pm$0.34 & 94.51$\pm$0.21 & -
  \\ \cmidrule(r){1-6}
  KFBNet (VGG16)~\cite{KFB} & 94.27$\pm$0.02 & 97.19$\pm$0.07 & 90.27$\pm$0.02 & 92.54$\pm$0.03 & $\textbf{99.76}\pm\textbf{0.24}$
  \\ \cmidrule(r){1-6}
  KFBNet (DenseNet121)~\cite{KFB} & 95.50$\pm$0.27 & 97.40$\pm$0.10 & $\textbf{93.08}\pm\textbf{0.14} $ & 95.11$\pm$0.10 & $\textbf{99.88}\pm\textbf{0.12}$
  \\ \cmidrule(r){1-6}
  MGML-FENet (VGG16) & $\textbf{94.47}\pm\textbf{0.15}$ & $\textbf{97.89}\pm\textbf{0.07}$ & 90.69$\pm$0.14 & $\textbf{93.36}\pm\textbf{0.12}$ & $\textbf{99.81}\pm\textbf{0.10}$
  \\ \cmidrule(r){1-6}
  MGML-FENet (ResNet34) & $\textbf{95.85}\pm\textbf{0.13}$ & $\textbf{98.44}\pm\textbf{0.06}$ &  91.39$\pm$0.18 & 94.54$\pm$0.07 & $\textbf{99.81}\pm\textbf{0.10}$
  \\ \cmidrule(r){1-6}
  MGML-FENet (DenseNet121) & $\textbf{96.45}\pm\textbf{0.18}$ & $\textbf{98.60}\pm\textbf{0.04}$ & 92.91$\pm$0.22 & $\textbf{95.39}\pm\textbf{0.08}$ & $\textbf{99.86}\pm\textbf{0.12}$
  \\ \cmidrule(r){1-6}
\end{tabular}
\end{center}
\end{table*}

\subsection{Ablation Study}
\par In our proposed models, we adopt different modules according to different motivations. To separately show the effectiveness of each module, we make more ablation experiments. In this section, we run all experiments on AID and NWPU-RESISC45 datasets.
\subsubsection{Comparison with baseline models}
In RS scene classification task, some notable deep convolutional neural networks can individually work well. Besides comparing with previous SOTA methods to show the effectiveness of our proposed method, we also compare results with baseline models' results. In this paper, we use VGG16, ResNet34 and DenseNet121 as baseline models. Tab.\ref{tab1} shows the detailed structure of them.
\par
\begin{figure}[!t]
  \centering
  \includegraphics[width=1.0\linewidth]{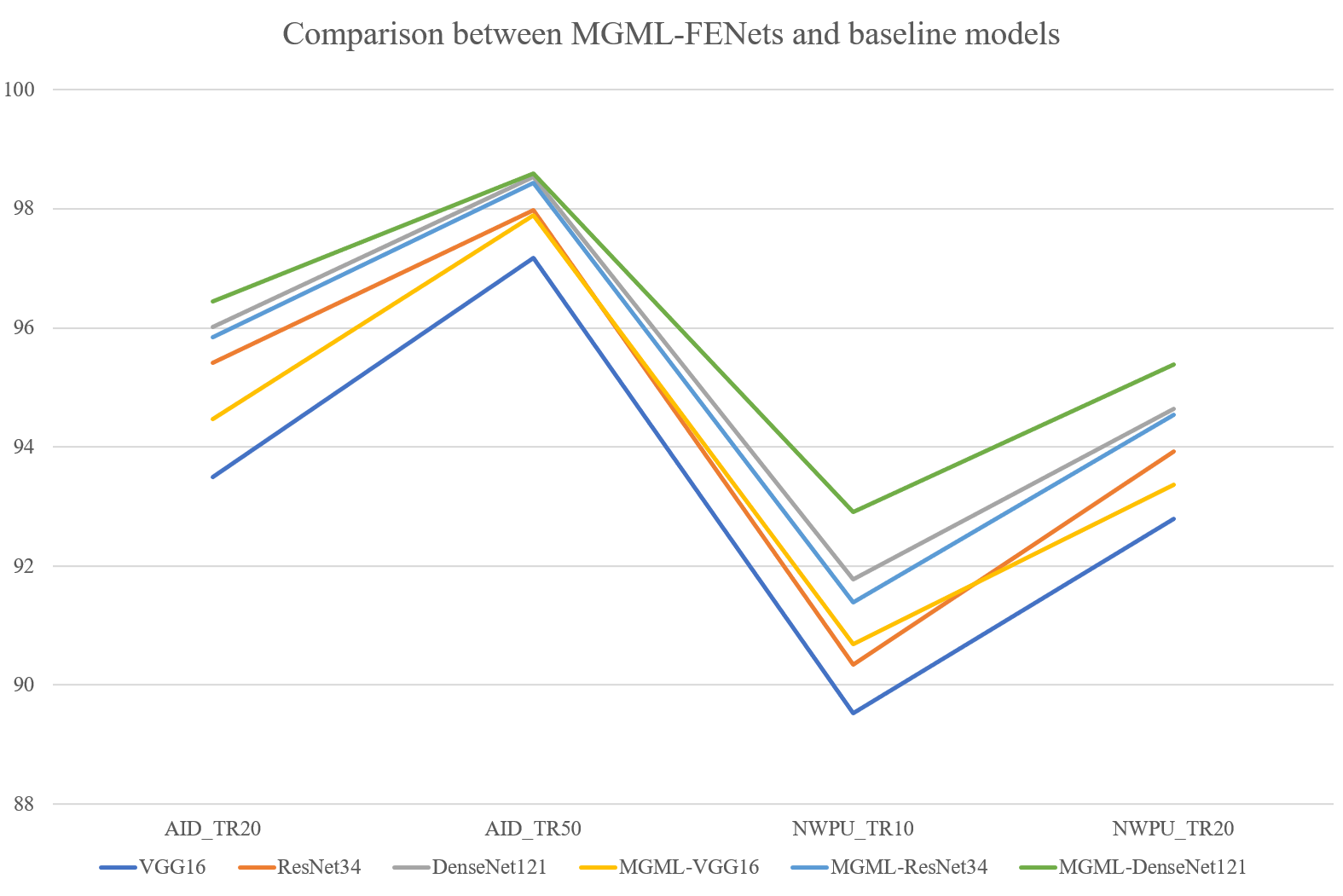}
  \caption{The comparison between MGML-FENets and baseline models. The curve shows the OA performance of our proposed models and baseline models on AID and NWPU-RESISC45 datasets with different training rate.}
\label{Fig5}
\end{figure}

 The comparison results between baseline models and MGML-FENets are shown in Fig.\ref{Fig5} and Tab.\ref{tab3}. On AID and NWPU datasets, MGML-FENets achieve better results obviously. Especially taking VGG16 as baseline model, MGML-FENet improves by large margin. On AID dataset, MGML-FENet respectively improves 0.98\% and 0.82\% than VGG16. On NWPU-RESISC45 dataset, MGML-FENet achieves 1.16\% and 0.57\% higher accuracy than VGG16. Based on ResNet34, MGML-FENet still has large improvement. Especially on NWPU-RESISC45 when the training rate is 10\%, our proposed model obtains 1.04\% (90.35\% $\sim$ 91.39\%). When the baseline model is DenseNet121, the classification results have already achieved high level. MGML-FENet further gains improvement. On NWPU-RESISC45, the leading gap is respectively 0.83\% and 0.65\%. Moreover, when using smaller group of training samples, MGML-FENets perform much better, which shows the robustness and effectiveness of our method.

\subsubsection{The effect of MGML-FFB and MGML-FEM}
To show the separate effect of MGML-FFB, we only apply main branch and MGML-FFB to form the whole network. Fig.\ref{Fig3} shows that the network will only have two predictions $\bm{P_{mb}}$ and $\bm{P_{ffb}}$ when removing MGML-FEM. From Tab.\ref{tab3}, we observe that, the mean OA of networks improves when adding MGML-FFB into baseline model. However, the standard deviation becomes bigger. The bigger fluctuation of results is because two branches extract different features and the predictions always tend to provide different votes for final results. Actually, adding MGML-FFB makes a trade-off between the advantage of diverse predictions and the fluctuation of negative votes.
\par MGML-FEM is designed to extract the high-level structural features. To show the effect of this module, we directly add MGML-FEM to baseline model and evaluate the classification performance. As shown in Tab.\ref{tab3}, compared to baseline models, networks only adding MGML-FEM have strong and stable performance with higher mean OA and lower standard deviation.
\begin{table}
\begin{center}
\caption{Ablation comparison experiments of classification results (\%) on AID and NWPU-RESISC45 datasets.}
\label{tab3}
\scalebox{0.8}{
\begin{tabular}{ccccc}
  \cmidrule(r){1-5}
  \multirow{2}{*}{Methods} & \multicolumn{2}{c}{AID} & \multicolumn{2}{c}{NWPU-RESISC45}
  \\ \cmidrule(r){2-5}
  {} & T.R.=20\% & T.R.=50\% & T.R.=10\% & T.R.=20\%
  \\ \cmidrule(r){1-5}
 VGG16 & 93.49$\pm$0.21 & 97.17$\pm$0.19 & 89.53$\pm$0.22 & 92.79$\pm$0.17
  \\ \cmidrule(r){1-5}
 VGG16+MGML-FFB & 93.71$\pm$0.27 & 97.62$\pm$0.15 & 89.88$\pm$0.30 & 92.92$\pm$0.28
 \\ \cmidrule(r){1-5}
 VGG16+MGML-FEM & 93.88$\pm$0.13 & 97.57$\pm$0.08 & 89.97$\pm$0.18 & 93.11$\pm$0.14
 \\ \cmidrule(r){1-5}
 MGML-FENet(VGG16) & 94.47$\pm$0.15 & 97.89$\pm$0.07 & 90.69$\pm$0.14 & 93.36$\pm$0.12
 \\ \cmidrule(r){1-5}
 ResNet34 & 95.42$\pm$0.17 & 97.98$\pm$0.11 & 90.35$\pm$0.27 & 93.92$\pm$0.15
  \\ \cmidrule(r){1-5}
 ResNet34+MGML-FFB & 95.73$\pm$0.24 & 98.00$\pm$0.12 & 90.54$\pm$0.2 & 94.14$\pm$0.16
 \\ \cmidrule(r){1-5}
 ResNet34+MGML-FEM & 95.74$\pm$0.13 & 98.26$\pm$0.08 & 90.96$\pm$0.16 & 94.33$\pm$0.12
  \\ \cmidrule(r){1-5}
 MGML-FENet(ResNet34) & 95.85$\pm$0.13 & 98.44$\pm$0.06 & 91.39$\pm$0.18 & 94.54$\pm$0.07
  \\ \cmidrule(r){1-5}
  DenseNet121 & 96.01$\pm$0.18 & 98.54$\pm$0.08 & 92.08$\pm$0.17 & 94.74$\pm$0.07
  \\ \cmidrule(r){1-5}
  DenseNet121+MGML-FFB & 96.14$\pm$0.28 & 98.56$\pm$0.15 & 92.23$\pm$0.23 & 94.89$\pm$0.13
  \\ \cmidrule(r){1-5}
  DenseNet121+MGML-FEM & 96.31$\pm$0.15 & 98.56$\pm$0.06 & 92.40$\pm$0.16 & 95.00$\pm$0.09
  \\ \cmidrule(r){1-5}
  MGML-FENet (DenseNet121) & 96.45$\pm$0.18 & 98.60$\pm$0.04 & 92.91$\pm$0.22 & 95.39$\pm$0.08
  \\ \cmidrule(r){1-5}
\end{tabular}}
\end{center}
\end{table}
\subsubsection{The effect of feature ensemble network}
Our proposed MGML-FENet is constructed by integrating main branch (baseline model) MGML-FFB and MGML-FEM together. Tab.\ref{tab3} shows clear that integrating MGML-FFB and MGML-FEM can gain better OA than applying each of them singly. With ensemble learning strategy, the whole network utilizes four predictions to vote for final results. And different branches provide predictions containing different features. Specifically, main branch focuses on extracting global feature. MGML-FFB extracts multi-granularity feature at different levels of network. MGML-FEM aims to utilize the structural information on high-level features. With feature ensemble learning strategy, MGML-FENets perform much stronger and stabler.
\subsubsection{7-crop vs 9-crop}
In this paper, we mainly adopt 7-crop both in RPM of CS-FG and FC-FG. Because we find the typical feature of RS images always appear in ``band'' areas (band in middle row and band in middle column) based on observation. Compared to 7-crop method, 9-crop method is another region proposal method which is more flexible. According to Alg.\ref{Alg1}, 9-crop can be easily expanded to ``$(k+1)^{2}$''-crop with the setting of different $s_{H}$ and $s_{W}$.
\par To compare the performance of 7-crop and 9-crop, we apply these two region proposal approaches respectively on MGML-FENets and keep other settings unchanged. The comparison results on AID and NWPU-RESISC45 datasets are shown in Tab.\ref{tab4}. Although 9-crop shows little weaker performance against 7-crop, It still has advantage on flexibility and extensibility.

\begin{table}
\begin{center}
\caption{Ablation comparison experiments of classification results (\%) on AID and NWPU-RESISC45 datasets.}
\label{tab4}
\begin{tabular}{ccc}
  \cmidrule(r){1-3}
  \multirow{2}{*}{Methods} & \multirow{2}{*}{\shortstack{AID \\ T.R.=20}} & \multirow{2}{*}{\shortstack{NWPU-RESISC45 \\ T.R.=10}}
  \\ \\ \cmidrule(r){1-3}
 MGML-FENet(VGG16)-7crop & 94.47$\pm$0.15 & 90.69$\pm$0.14
 \\ \cmidrule(r){1-3}
 MGML-FENet(VGG16)-9crop & 94.29$\pm$0.14 & 90.71$\pm$0.17
 \\ \cmidrule(r){1-3}
 MGML-FENet(ResNet34)-7crop & 95.85$\pm$0.13 & 91.39$\pm$0.18
  \\ \cmidrule(r){1-3}
  MGML-FENet(ResNet34)-9crop & 95.80$\pm$0.10 & 91.22$\pm$0.13
  \\ \cmidrule(r){1-3}
  MGML-FENet (DenseNet121)-7crop & 96.45$\pm$0.18 & 92.91$\pm$0.22
  \\ \cmidrule(r){1-3}
  MGML-FENet (DenseNet121)-9crop & 96.31$\pm$0.21 & 92.83$\pm$0.16
  \\ \cmidrule(r){1-3}
\end{tabular}
\end{center}
\end{table}

\subsection{Visualization and Analysis}
\subsubsection{Convergence analysis}
\begin{figure}[!t]
  \centering
  \includegraphics[width=1.0\linewidth]{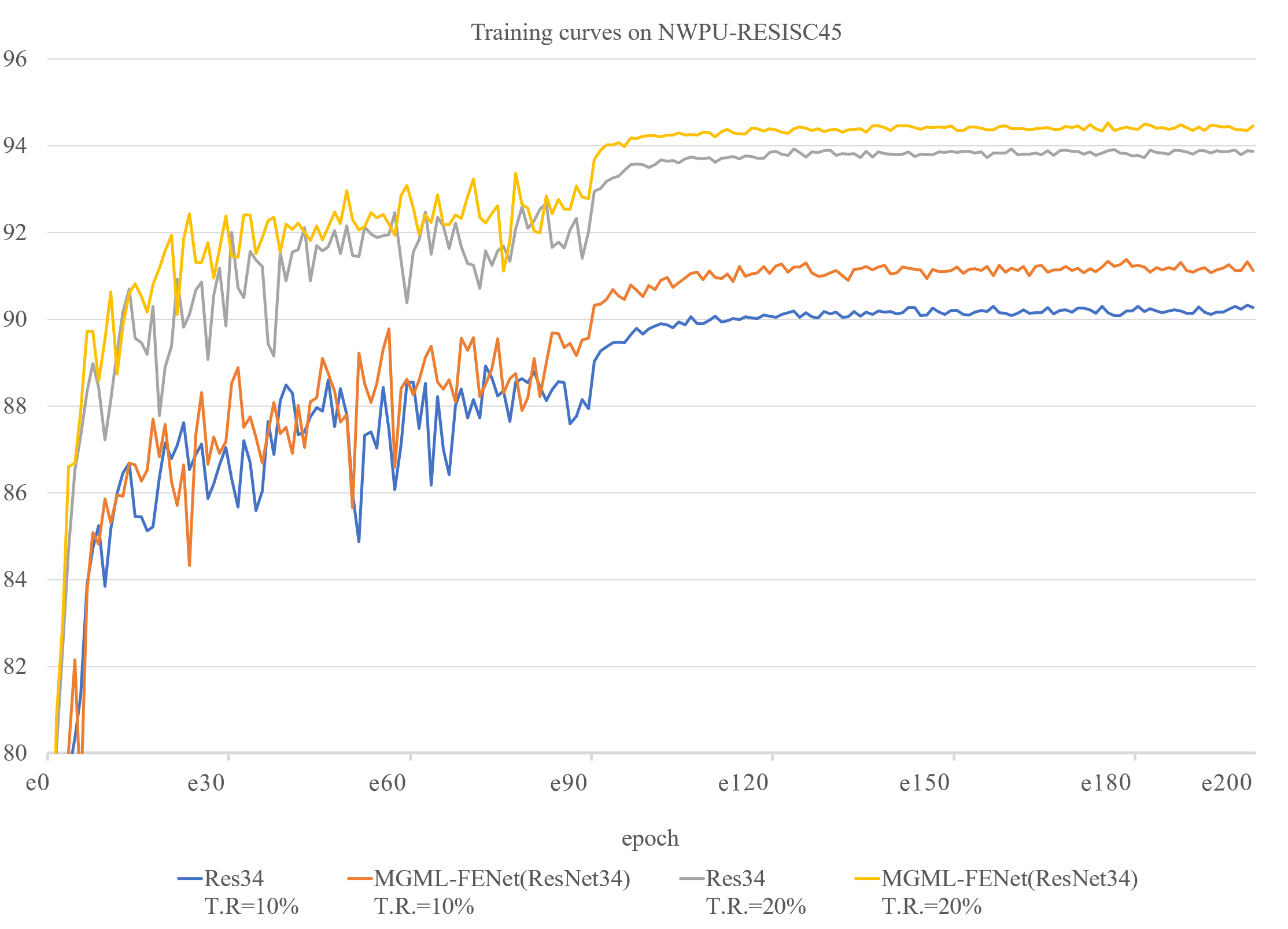}
  \caption{The OA(\%)/epoch training curves of MGML-FENet (ResNet34) and ResNet34 on NWPU-RESISC45. }
\label{Fig6}
\end{figure}

Training MGML-FENets aims to optimize objective functions $L_{obj}$. In Fig.\ref{Fig6}, we select ResNet34 as baseline model and use the classification results on NWPU-RESISC45 as an example to analyze the convergence by showing the ``OA-epoch'' curves. As shown in Fig.\ref{Fig6}, MGML-FENets can converge smoothly even with more complex objective functions to optimize. Moreover, MGML-FENets obviously has higher overall accuracy than baseline model (ResNet34) after converging.

\begin{figure*}[!t]
  \centering
  \includegraphics[width=1.0\linewidth]{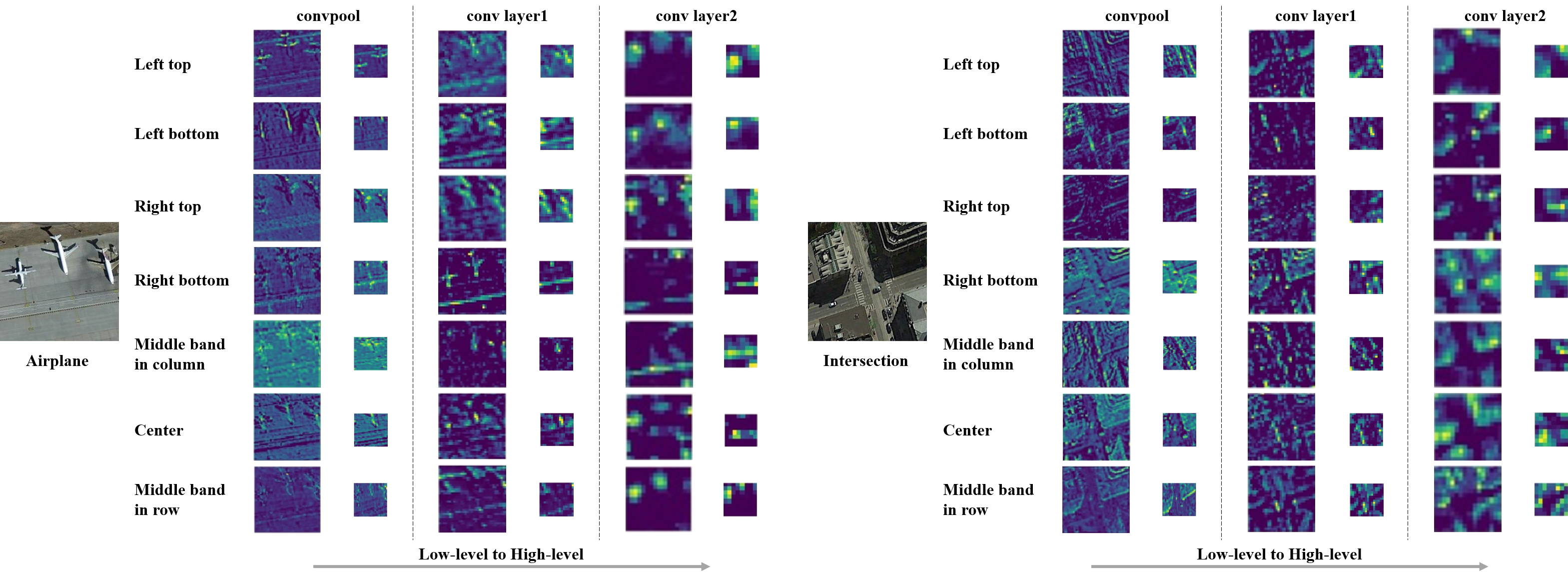}
  \caption{Feature map visualization of MGML-FENet (ResNet34) on NWPU-RESISC45. The two images are randomly selected during testing and used to generate feature maps. The feature maps are selected from different levels of networks. In each feature map pair, the left feature map global feature map of main branch ($\bm{G_{i}}$) and the right feature map ($\bm{H_{i}}$) is cropped and pooled by the left feature map through CS-FG. For different local feature patches (7-crop), we randomly select one-channel feature map to visualize. The output feature maps are respectively from ``conv1 poo1'', ``conv layer1'' and ``conv layer2''. Moreover, the color in feature map indicates the pixel intensity. The warmer the pixel colour, the bigger the pixel activation.}
\label{Fig7}
\end{figure*}

\subsubsection{Feature map visualization and analysis}
To intuitively interpret out proposed method, we visualize feature map in different levels of network. We select MGML-FENet (ResNet34) to run experiments on NWPU-RESISC45 with T.R.=20\%. When the model converges, we visualize feature maps to observe the attention area. From Fig.\ref{Fig7}, we analyze our proposed method in the following five points.
\par First, CS-FG can extract multi-granularity features to help reduce negative influence from large intra-class variance. Following the explanation of~\cite{MGCAP}, the global feature map ($\bm{G_{i}}$) can be regarded as $1^{st}$ granularity feature. Through 7-crop region proposal module of CS-FG, the global feature map is cropped and pooled. The output feature patches can be seen to contain the characteristic of $2^{nd}$ granularity. When we concatenate feature patches together, the new feature maps ($\bm{H_{i}}$) both contain the separate features from different feature patches and the structural feature by combining different feature patches. If we regard the structural feature as $3^{rd}$ granularity feature, the output from CS-FG contain both $2^{nd}$ and $3^{rd}$ granularity feature. All in all, with main branch and MGML-FFB, MGML-FENets utilize multi-granularity feature to enhance the network performance.
\par Second, our proposed networks integrate feature maps at different levels which can improve generalization ability. As shown in Fig.\ref{Fig7}, feature maps at different level of networks contain different information. In MGML-FENets, MGML-FFB and MGML-FEM both extract and fuse different level feature maps.
\par Third, MGML-FENets can obtain abundant fine-grained features by CS-FG which can help network learn distinct characteristics of each category. For example, in the ``Airplane'' image, some features (Left top, Right top, $\cdots$) have attention on the planes. Planes are the most distinct character of category ``Airplane''. Besides planes, some feature patches (Right bottom, Middle band in row, $\cdots$) focus on the runway which is also significant character to recognize category ``Airplane''. In RS images, planes in ``Airplane'' images are sometimes very small. Under this situation, other fine-grained features like runway will make a big difference for classification.
\par Fourth, RS images has large resolution and wide cover range. Extracting local patches can help network filter redundant and confusing information. In Fig.\ref{Fig7}, it is apparently that the attention region in some feature patches become clearer (color become warmer) than in global feature map. For example, in the ``Intersection'' image, the feature maps usually have equally attention intensity on the edges of roads or road corners which will lower the contrast. Using local feature patches can enhance the attention intensity in different local regions. E.g. The ``right bottom'' patches will only focus on the edge information of right bottom road corner and the ``middle band in column'' will focus on the edge information of horizontal road. All in all, Extracting local patches can enhance attention intensity and get enhanced fine-grained features through adaptive pooling on smaller local patches with less interference.
\par Last but not least, channel-separate strategy can guide global feature maps to have different focuses. Because of this, the networks become compact and efficient. Specifically, channel-separate strategy forces the networks to recognize through a group of local feature patches. And only few channels are provided for each local patch. Through experiments and visualization (Fig.\ref{Fig7}), we find that global feature maps tend to have similar attention regions and patterns with corresponding feature patches. It is positive because abundant feature representation can improve the performance of networks.

\subsubsection{Predictions visualization analysis with T-SNE}
\begin{figure*}[!t]
  \centering
  \includegraphics[width=1.0\linewidth]{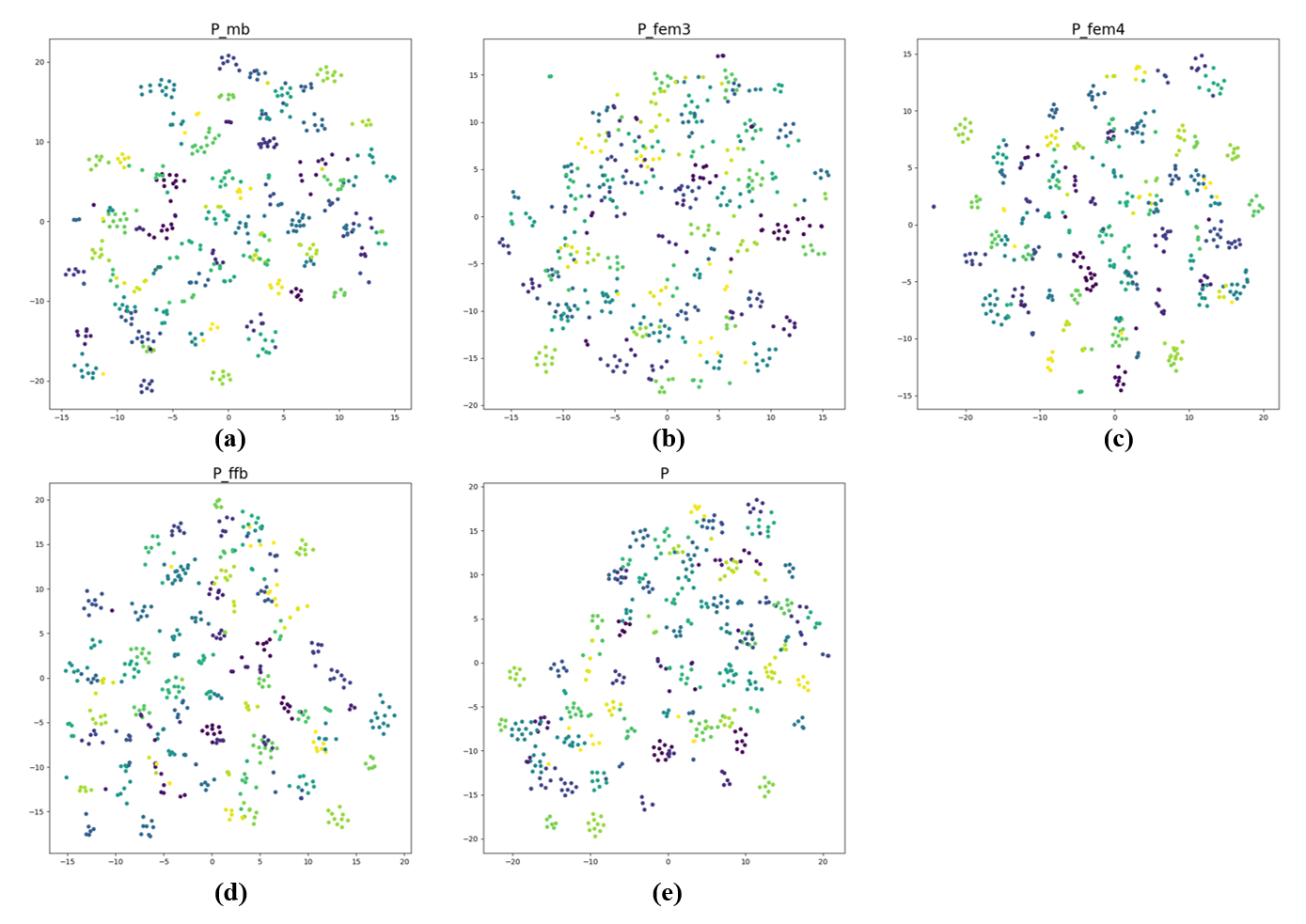}
  \caption{Visualization results of four predictions of MGML-FENet. We reduce 45-dimension prediction vectors to 2-dimension by T-SNE. Additionally, we use MGML-FENet(ResNet34) on NWPU-RESISC45 and randomly select 512 testing samples to visualize.}
\label{Fig8}
\end{figure*}

Inspired by ensemble learning method, We assume that the final voting accuracy will become higher if the four predictions can provide diverse and accurate results. To intuitively show the distribution patterns of four predictions, we apply T-SNE~\cite{T_SNE} method to visualize and analyze $\bm{P_{mb}}, \bm{P_{ffb}}, \bm{P_{fem3}}, \bm{P_{fem4}}$ and $\bm{P}$. The visualization results are shown in Fig.\ref{Fig8}.
\par From Fig.\ref{Fig8}, we analyze in the following three points. First, the four predictions all have reasonable classification results on 45 categories. Even though some samples are still confusing and hard to classify, the category clusters are clear. Second, cluster maps of the four predictions have diverse patterns which is helpful for the network to deal with confusing samples. Third, the final predictions ($\bm{P}$) have better cluster feature distribution. Obviously, points in clusters are tighter (smaller intra-class distance) and distance between clusters are larger (larger inter-class distance). All in all, Fig.\ref{Fig8} proves the effectiveness and interpretability of our feature ensemble network.

\subsubsection{Computation cost analysis}
Compared to baseline models, MGML-FENets have more computation cost during inference time. In MGML-FFB, more ``conv layers'' are introduced which cause more convolution operation. However, in MGML-FFB, feature maps in each level of networks are cropped into several feature patches and recombined together by CS-FG. New feature maps have equal channels but less spatial scale as original feature maps. Therefore, the computation increment are restrained. We list the computation cost comparison In Tab.\ref{tab5}.
\par MGML-FENets have more computation cost than baseline models. In (Tab.\ref{tab2}), MGML-FENets earn accuracy improvement by big margin (more than 1\% in some cases), even though some extra inference computation are introduced. In practical application, we always need to control computation cost. Therefore, ``baseline+MGML-FEM'' networks are more efficient choices. From Tab.\ref{tab2} and Tab.\ref{tab5}, we know that ``baseline+MGML-FEM'' networks can gain average 0.4 $\sim$ 0.5\% OA improvement with almost no extra computation costs.

\begin{table}
\begin{center}
\caption{Comparison of computation costs (FLOPs). The input image scale is set as 224*224.}
\label{tab5}
\begin{tabular}{cc}
  \cmidrule(r){1-2}
 Methods & FLOPs
  \\ \cmidrule(r){1-2}
 VGG16 & 15.4G
  \\ \cmidrule(r){1-2}
 VGG16+MGML-FEM & 15.4G
 \\ \cmidrule(r){1-2}
 MGML-FENet(VGG16) & 36.5G
 \\ \cmidrule(r){1-2}
 ResNet34  & 3.6G
  \\ \cmidrule(r){1-2}
 ResNet34+MGML-FEM & 3.6G
  \\ \cmidrule(r){1-2}
 MGML-FENet(ResNet34) & 4.8G
  \\ \cmidrule(r){1-2}
  DenseNet121 &  2.9G
  \\ \cmidrule(r){1-2}
  DenseNet121+MGML-FEM & 2.9G
  \\ \cmidrule(r){1-2}
  MGML-FENet (DenseNet121) & 4.4G
  \\ \cmidrule(r){1-2}
\end{tabular}
\end{center}
\end{table}

\section{Conclusion}
In this paper, we design a multi-granularity multi-level feature ensemble network to tackle RS scene classification task. In MGML-FENet, Main branch is used for maintain useful global feature. MGML-FFB is employed to extract multi-granularity feature and explore fine-grained features in different levels of networks. MGML-FEM is designed to utilize high-level features with structural information. Specifically, we propose two important module: channel-separate feature generator and full-channel feature generator to extract feature patches and recombine them. Extensive experiments show that the proposed networks outperform the previous models and achieve SOTA results on notable benchmark datasets in RS scene classification task. In addition, visualization results prove that our proposed networks are reasonable and interpretable.


%
\appendices
\section{Classification Results on VGoogle dataset}
Compared to AID, NWPU and UC-Merced, VGoogle is a new RS dataset containing more samples. We evaluate our method on VGoogle to further show the general performance of MGML-FENets. We select ResNet34 as baseline model and run experiments with 5\% and 10\% training rate. We report our encouraging comparison results in Tab.\ref{tab6}.  When using low training rate (5\%), MGML-FENet(ResNet34) performs obvious better with 0.77\% OA improvement. When the training rate is set to 10\%, MGML-FENet(ResNet34) also achieve better OA. Results on VGoogle prove that out proposed MGML-FENets have convincing general performance.

\begin{table}
\begin{center}
\caption{Comparison experiments of classification results (\%) on VGoogle dataset between baseline model (ResNet34) and MGML-FENets.}
\label{tab6}

\begin{tabular}{ccc}
  \cmidrule(r){1-3}
  \multirow{2}{*}{Methods} & \multicolumn{2}{c}{VGoogle}
  \\ \cmidrule(r){2-3}
  {} & T.R.=5\% & T.R.=10\%
  \\ \cmidrule(r){1-3}
 ResNet34 & 96.37$\pm$0.18 & 97.81$\pm$0.10
  \\ \cmidrule(r){1-3}
 MGML-FENet(ResNet34) & 97.14$\pm$0.16 & 98.17$\pm$0.09
 \\ \cmidrule(r){1-3}
\end{tabular}
\end{center}
\end{table}

\section*{Acknowledgment}
This work was supported by the National Natural Science Foundation of China [grant number 62072021]

\ifCLASSOPTIONcaptionsoff
  \newpage
\fi



%

\bibliographystyle{IEEEtran}
\bibliography{IEEEfull,mybibfile}

%

\begin{IEEEbiography}[{\includegraphics[width=1in,height=1.25in,clip,keepaspectratio]{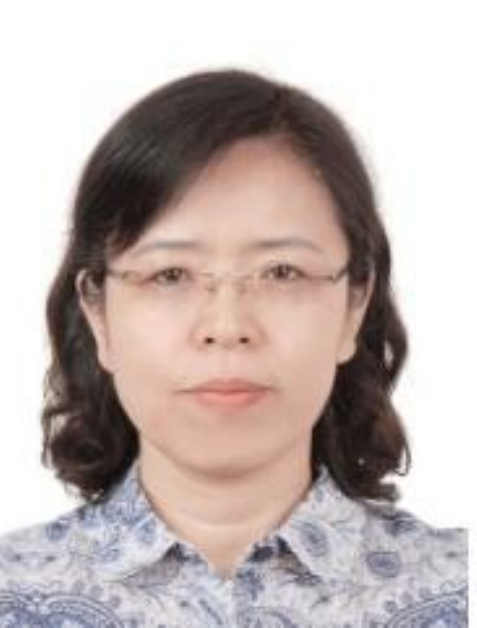}}]{Qi Zhao}
Qi Zhao received Ph.D in communication and information system from Beihang University, Beijing, China, in 2002. From 2002 to date, she is an associate professor and works in Beihang University. She was in the Depart-
ment of Electrical and Computer Engineering at the University of Pittsburgh as a visiting scholar from 2014 to 2015. Her current research interests include image recognition and processing, communication signal processing and target tracking.
\end{IEEEbiography}
\begin{IEEEbiography}[{\includegraphics[width=1in,height=1.25in,clip,keepaspectratio]{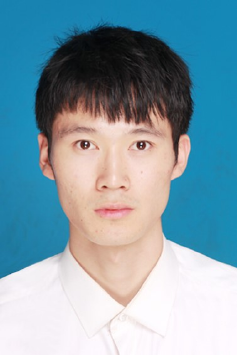}}]{Shuchang Lyu} received the B.S. degree in communication and information from Shanghai University, Shanghai, China, in 2016, and the M.E. degree in communication and information system from the School of Electronic and Information Engineering, Beihang University, Beijing, China, in 2019. He is currently pursuing the Ph.D. degree with the School of Electronic and Information Engineering, Beihang University, Beijing. His research interests include deep learning, image classification, one-shot semantic segmentation and object detection.
\end{IEEEbiography}
\begin{IEEEbiography}[{\includegraphics[width=1in,height=1.25in,clip,keepaspectratio]{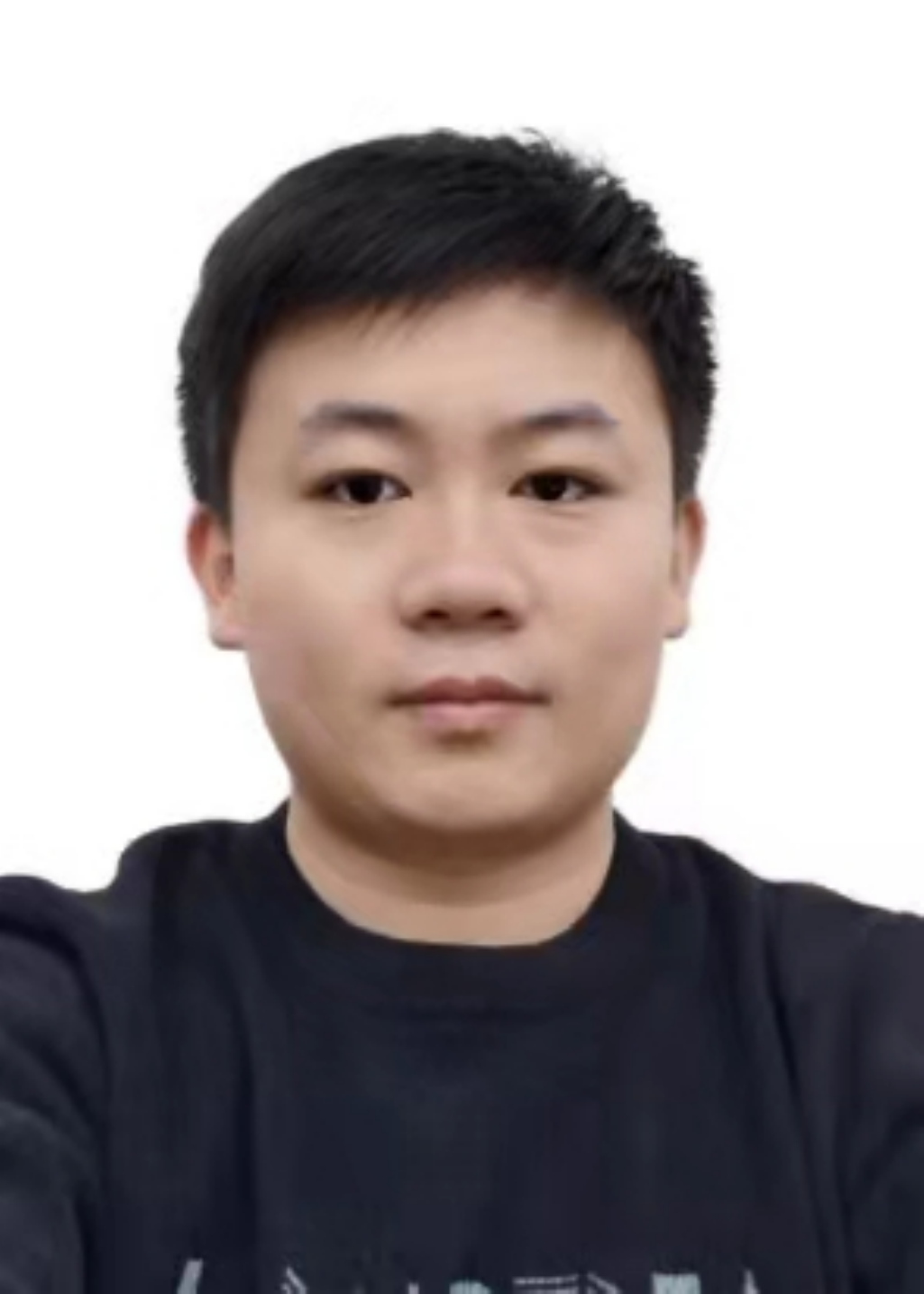}}]{Yuewen Li} received B.S. degree in electronics and
information engineering from Beihang University,
Beijing, China, in 2017. He is currently pursuing the
M.S. degree with the school of electronics and information engineering, Beihang University, Beijing,
China. His research interests include deep learning,
object detection and few-shot learning.
\end{IEEEbiography}
\begin{IEEEbiography}[{\includegraphics[width=1in,height=1.25in,clip,keepaspectratio]{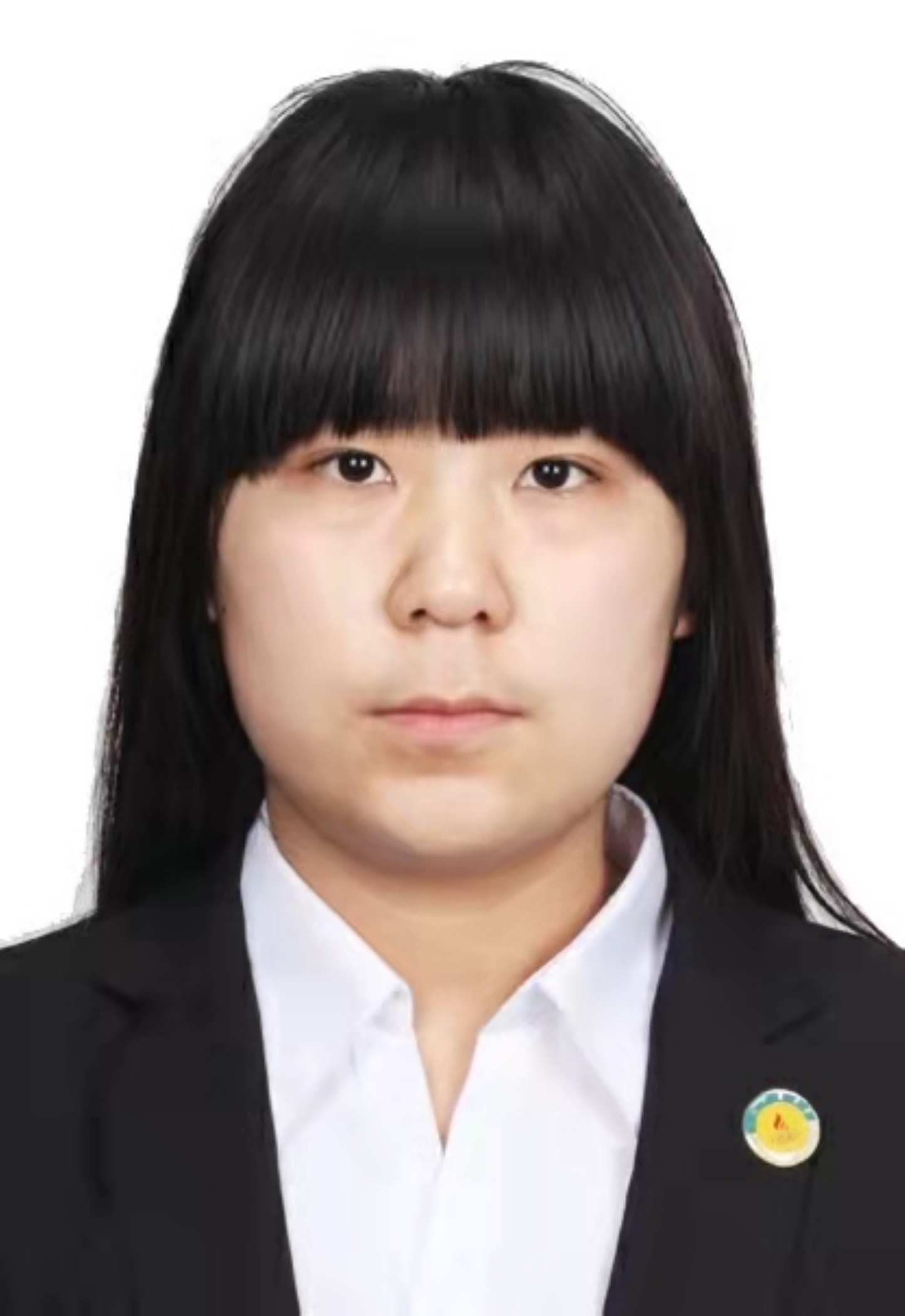}}]{Yujing Ma} received B.S. degree in electronics and
information engineering from Beihang University,
Beijing, China, in 2017. She is currently pursuing the M.S. degree with the school of electronics and information engineering, Beihang University, Beijing,
China. Her research interests include deep learning,
Remote sensing and active learning.
\end{IEEEbiography}
\begin{IEEEbiography}[{\includegraphics[width=1in,height=1.25in,clip,keepaspectratio]{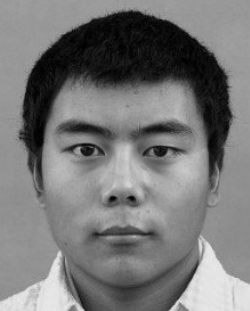}}]{Lijiang Chen}
received B.S. and Ph.D. degrees in the School of Electronic and Information Engineering from Beihang University in 2007 and 2012 respectively. He was a Hong Kong Scholar in the City University of Hong Kong from 2015 to 2017. He is now an assistant professor with the School of Electronic and Information Engineering, Beihang University. His current research interests include pattern recognition, image processing, and human-computer interaction.
\end{IEEEbiography}





\end{document}